\documentclass[twoside]{article}

\usepackage[accepted]{aistats2023}

\usepackage[round]{natbib}

\usepackage{mathtools}
\usepackage{amssymb}
\usepackage{bm}

\usepackage{algorithm}
\usepackage{algorithmic}

\usepackage{multirow}
\usepackage{booktabs}
\usepackage{tikz}
\usepackage{url}
\usepackage{hyperref}
\usepackage{cleveref}

\usepackage{textcomp}

\DeclareMathOperator{\dims}{dims}
\DeclareMathOperator{\OHE}{OHE}

\newcommand{\R}{\mathbb{R}}
\newcommand{\E}{\mathbb{E}}

\newtheorem{lemma}{Lemma}
\newenvironment{proof}
{\par\noindent{\bf Proof.}} 
{\hfill$\scriptstyle\blacksquare$}

\begin{document}

\twocolumn[

\aistatstitle{MARS: Masked Automatic Ranks Selection in Tensor Decompositions}

\aistatsauthor{ Maxim Kodryan \And Dmitry Kropotov \And Dmitry Vetrov }

\aistatsaddress{ HSE University \And  Lomonosov Moscow State University \\ HSE University \And HSE University \\ AIRI } ]

\begin{abstract}
  Tensor decomposition methods have proven effective in various applications, including compression and acceleration of neural networks. 
  At the same time, the problem of determining optimal decomposition ranks, which present the crucial parameter controlling the compression-accuracy trade-off, is still acute. 
  In this paper, we introduce MARS~--- a new efficient method for the automatic selection of ranks in general tensor decompositions. 
  During training, the procedure learns binary masks over decomposition cores that ``select'' the optimal tensor structure. 
  The learning is performed via relaxed maximum a posteriori (MAP) estimation in a specific Bayesian model and can be naturally embedded into the standard neural network training routine.
  Diverse experiments demonstrate that MARS achieves better results compared to previous works in various tasks.
\end{abstract}

\section{INTRODUCTION}

Tensor decomposition methods are leveraged in various areas of machine learning, such as multi-way data analysis~\citep{cichocki2009nonnegative}, higher-order representation learning~\citep{castellana2019bayesian}, recommender systems~\citep{frolov2017tensor}, and many others~\citep{ji2019survey}. 
Perhaps the most famous and perspective application of these techniques is deep neural networks (DNNs) tensorization~\citep{cichocki2017tensor}. 
Decomposition methods cope with redundancy in DNNs parameterization via an efficient representation of neural network parameters as decomposed tensors. 
Recent works on applying tensor decomposition techniques in neural networks have demonstrated the success of this approach for compression, speed-up, and regularization of DNN models.
For instance, Tucker~\citep{Tucker1966} and canonical polyadic (CP)~\citep{carroll1970analysis} tensor decompositions are widely used for compressing and accelerating convolutional neural networks~\citep{DBLP:journals/corr/LebedevGROL14, DBLP:journals/corr/KimPYCYS15, kossaifi2019t}, and Tensor Train (TT)~\citep{TT:Oseledets:11} decomposition has been successfully applied for tensorization of a variety of neural networks layers, like fully-connected (FC)~\citep{novikov15tensornet}, convolutional~\citep{garipov16ttconv}, recurrent~\citep{DBLP:conf/icml/YangKT17, yu2017long}, embedding~\citep{hrinchuk2020tensorized}, etc.
 
Probably the main crux of the tensorization approach is the need to carefully select decomposition hyperparameters, namely the ranks.
Tensor decomposition ranks determine the shape of the core tensors in the decomposition (the cores) and hence are responsible for the trade-off between the quality of the model and the required computational and memory resources.
Therefore, decomposition ranks represent extremely important hyperparameters.
Yet, the problem of finding an efficient way to select optimal ranks in a general tensor decomposition automatically still remains unresolved.
Typical hyperparameter selection techniques, like cross-validation, are poorly suited for the choice of multiple tensor ranks.
Hence, the common practice is to set all ranks equal and validate a single hyperparameter. 
However, such a simplification is quite coarse and might significantly degrade model performance compared to a non-uniform ranks selection, which we empirically demonstrate in our experiments.

In this work, we present \emph{Masked Automatic Ranks Selection} (MARS)~--- a new efficient method for dynamic selection of tensor decomposition ranks grounded in the Bayesian framework. 
The main idea is to learn binary masks that cover decomposition cores and ``select'' only the slices required for the best model performance, thus automatically adjusting the optimal ranks arrangement.
We also propose a way to reformulate the emerged NP-hard discrete problem via scalable continuous optimization. 
MARS operates end-to-end with model training without introducing any noticeable additional computational overhead.
In sum, our \textbf{major contributions} consist in 1) proposing a \emph{general} scheme for automatic ranks selection and 2) developing an \emph{efficient scalable} method for applying that scheme.

In the experiments, we demonstrate that our method is applicable for various tensorized neural networks (TNNs) and even more general tensorized models (\Cref{app:tucker_approx}).
We evaluate MARS on a variety of tasks and architectures involving convolutional, fully-connected, and embedding tensorized layers and demonstrate its ability to improve previous results on tensorization in terms of compression, accuracy, and speed-up.
Our code is available at \url{https://github.com/MaxBourdon/mars}.

\subsection{Related work}

Here we highlight related work on tensor rank determination with a focus on application to deep learning.
We refer to \Cref{app:rel_work} for more related literature on tensorization.

\citet{DBLP:journals/corr/KimPYCYS15} perform full DNN compression via approximating FC and convolutional layers with low-rank matrix factorization and Tucker-2 tensor decomposition, respectively, where ranks are estimated with a special Bayesian matrix rank selection technique~\citep{nakajima2012perfect}.
However, the involved training procedure consisting of decomposition of the pre-trained model and fine-tuning of the decomposed model turned out to be inefficient.
The MUSCO algorithm~\citep{gusak2019automated}, which repeatedly performs decomposition and fine-tuning steps, partially resolved this disadvantage.
Lately,~\citet{cheng2020novel} proposed a reinforcement learning-based rank selection scheme for TNNs, which, however, also introduces extra computational requirements by separating agent and model training.
In contrast, MARS operates end-to-end with model training without splitting it into stages, which is naturally more preferable. 
Moreover, it is not confined to specific types of tensor decompositions, models, or tasks.

Existing methods for automatic rank selection that also take advantage of the Bayesian approach cover only certain types of tensor decompositions~\citep{rai2014scalable, zhao2015bayesian, xu2020learning, xu2021probabilistic, fang2021bayesian} or are based on peculiarities of the task, e.g., tensor approximation~\citep{doi:10.1002/cem.1223} or linear regression~\citep{guhaniyogi2017bayesian}. 
These approaches mostly embody structured pruning of the decomposition cores.
For instance,~\citet{hawkins2021bayesian} propose a specific shrinking coupling prior distribution over TT-cores and perform Bayesian inference to obtain Low-Rank Bayesian Tensorized Neural Networks (LR-BTNN).
We, instead, propose a general-purpose ranks selection technique applicable for any tasks and decompositions.

Alternative procedures for obtaining low-rank tensor representation, e.g., those utilizing nuclear norm minimization~\citep{DBLP:journals/corr/PhienTBD16, DBLP:conf/nips/ImaizumiMH17, shi2021robust}, also leverage properties of the particular objective and/or suggest excessively computationally complex algorithms involving a series of SVDs.
This makes such approaches impracticable in the domain of deep learning. 
MARS does not impose any significant extra computations for obtaining a low-rank tensorized solution, since slice-wise mask multiplication is a much less computationally expensive operation than tensors contraction.

\section{MARS}
\label{sec:method}

In this section, we introduce the necessary notions regarding tensors, decompositions, and general tensorized models and describe the details of the proposed method.

\subsection{Tensors, decompositions, and tensorized models}

\paragraph{Tensors} 
By a \emph{$d$-dimensional tensor}, we mean a multidimensional array $\mathcal{A} \in \R^{n_1 \times \dots \times n_d}$ of real numbers, e.g., vectors and matrices are $1$- and $2$-dimensional tensors, respectively. 
We denote $\mathcal{A}(i_1, \dots, i_d)$ as element $(i_1, \dots, i_d)$ of a tensor $\mathcal{A}$. 
We use notation $\dims\left(\mathcal{A}\right) = (n_1, \dots, n_d)$ to denote the tuple of dimensions of a tensor $\mathcal{A}$.

\emph{Contraction} of two tensors $\mathcal{A} \in \R^{n_1 \times \dots \times n_d}$ and $\mathcal{B} \in \R^{m_1 \times \dots \times m_{d'}}$ with $n_d = m_1$ results in a tensor $\mathcal{A}\mathcal{B} \in \R^{n_1 \times \dots \times n_{d-1} \times {m_2} \dots \times m_{d'}}$:
\begin{multline}
    \label{eq:contraction}
    \mathcal{A}\mathcal{B}(i_1, \dots, i_{d-1}, j_2, \dots, j_{d'}) = \\
    =\sum_{i_d=1}^{n_d} \mathcal{A}(i_1, \dots, i_d) \mathcal{B}(i_d, j_2, \dots, j_{d'}).
\end{multline}
The contraction operation can also be naturally generalized to multiple modes. 
In this case, summation in \cref{eq:contraction} is performed over these modes, and dimensions of the resulting tensor will contain the dimensions of both tensors $\mathcal{A}$ and $\mathcal{B}$ excluding the contracted ones.

A special case of contraction (up to modes permutation) for a tensor $\mathcal{A} \in \R^{n_1 \times \dots \times n_d}$ and a matrix $B \in \R^{m_k \times n_k}$ is their \emph{mode-$k$ product} $\mathcal{A} \times_k B \in \R^{n_1 \times \dots \times n_{k-1} \times m_k \times n_{k+1} \times \dots \times n_d}$:
\begin{multline*}
    \left(\mathcal{A} \times_k B\right)(i_1, \dots, i_{k-1}, j_k, i_{k+1}, \dots, i_d) = \\
    =\sum_{i_k=1}^{n_k} \mathcal{A}(i_1, \dots, i_d) \mathcal{B}(j_k, i_k).
\end{multline*}

We also introduce \emph{mode-$k$ broadcast Hadamard product} of a tensor $\mathcal{A} \in \R^{n_1 \times \dots \times n_d}$ and a vector $b \in \R^{n_k}$ which is a tensor $\mathcal{A} \odot_k b$ with the same dimensions as $\mathcal{A}$ and elements
\begin{equation*}
    \left(\mathcal{A} \odot_k b\right)(i_1, \dots, i_d) = \mathcal{A}(i_1, \dots, i_d) b(i_k).
\end{equation*}

\paragraph{Tensor decompositions} 
In general, we assume that tensor decomposition of a $d$-dimensional tensor $\mathcal{A}$ consists of a set of simpler tensors $\bm{G} = \left\{\mathcal{G}_k\right\}$ called \emph{cores} of the decomposition. 
The original tensor can be expressed (up to modes permutation) via these cores as a sequence of contractions. 

For the Tensor Train decomposition $\bm{G} = \left\{\mathcal{G}_1, \dots, \mathcal{G}_d\right\}$, $\mathcal{G}_k \in \R^{r_{k-1} \times n_k \times r_k}$, $r_0 = r_d = 1$ and
\begin{equation*}
    \mathcal{A} = \mathcal{G}_1 \mathcal{G}_2 \dots \mathcal{G}_d,
\end{equation*}
i.e., tensor $\mathcal{A}$ is directly obtained from the Tensor Train cores as a sequence of contractions.

For the Tucker decomposition $\bm{G} = \left\{\mathcal{G}, U_1, \dots, U_d\right\}$, $U_k \in \R^{n_k \times r_k}$, $\mathcal{G} \in \R^{r_1 \times \dots \times r_d}$ and
\begin{equation*}
    \mathcal{A} = \mathcal{G} \times_1 U_1 \dots \times_d U_d,
\end{equation*}
i.e., tensor $\mathcal{A}$ is expressed via mode-$k$ products of the core tensor $\mathcal{G}$ and matrices $U_k$ which is again a sequence of contractions up to modes permutation.

The set of numbers $\bm{r} = \{r_k\}$, the intermediate dimensions of the contracted cores modes, are called \emph{ranks} of the decomposition. 
Clearly, they define the expressivity of the decomposition on the one hand and the number of the occupied parameters on the other. 

\paragraph{Tensorized models}
Consider any model parameterized by a tensor $\mathcal{A}$ decomposed into cores $\bm{G}$.\footnote{
For simplicity, we consider a single-tensor model, though the same applies to models with multiple tensors.} 
In practice, it is often convenient (in terms of memory and computations) to utilize tensors in the decomposed format explicitly. 
In other words, given a particular decomposition, one could rewrite model operations more efficiently via the cores $\bm{G}$ directly, without the need of reconstructing the full tensor~$\mathcal{A}$.
Hence, a single large parameter tensor can be substituted with a set of smaller tensors to obtain a more compact model. 
We call such models, parameterized by the cores of the decomposed tensors, \emph{tensorized models} and assume that their inference is performed directly via these cores. 

A typical example of a tensorized model is a neural network with decomposed layers, or tensorized neural network. 
Representing layer parameters via a decomposed tensor may result in substantial memory and computational savings.
For most NN layers, there is a variety of decomposed representations: factorized FC-layer, Tucker-2 convolutional layer, various TT-layers, etc. 
We provide more detail in \Cref{app:tens_layers}.

Ultimately, in a tensorized model, the shapes of the decomposition cores simultaneously influence model flexibility and complexity. 
The key hyperparameter that determines them are decomposition ranks, as discussed earlier. 
Further, we describe the details of the proposed method for ranks selection in arbitrary tensorized models.

\subsection{The proposed method}

Consider a predictive tensorized model, which defines a distribution over output $y$ conditioned on input $x$, with cores $\bm{G}$: $p\left(y \mid x, \bm{G}\right)$.
We assume that the initial shapes of the cores (i.e., ranks $\bm{r}$) are fixed in advance.
Our goal is to shrink them optimally: remove redundant ranks without significant accuracy drop to achieve maximum compression and speed-up.

MARS suggests obtaining such reduced structures via multiplying slices of the cores by binary masking vectors, mostly consisting of zeros.
Zeroed slices are not be involved in tensors contractions and, therefore, the whole model workflow.
Hence, such slices can be freely removed from the cores.
In this way, non-zero masks elements would ``select'' only slices required for the effectual model performance, automatically determining the optimal cores shapes.
\Cref{fig:mars_concept} illustrates the concept.

\begin{figure}[ht]
\centering
\includegraphics[width=0.95\linewidth]{MARS.png}
\caption{A schematic illustration of the MARS concept: slices of the core tensor $\mathcal{G}$ along mode $k$ are multiplied by elements of the binary mask $m_k$; only ``selected'' non-zero slices will participate in model inference, therefore, the core shape can be reduced.}
\label{fig:mars_concept}
\end{figure}

Formally, given a dataset $\left(X, Y\right) = \{(x_i, y_i)\}_{i=1}^N$, consider the following discriminative Bayesian model:\footnote{The presented model can be straightforwardly extended to other non-discriminative settings even with no available data points, see \Cref{app:tucker_approx} for a concrete example.}
\begin{equation}
    \label{eq:model}
    p\left(Y, \bm{m}, \bm{G} \mid X\right) = \prod_{i=1}^N p\left(y_i \mid x_i, \bm{G} \odot \bm{m}\right) p\left(\bm{m}\right) p\left(\bm{G}\right),
\end{equation}
where $\bm{m} = \{m_k \mid m_k \in \{0, 1\}^{r_k}\}$ is a set of binary vectors, or \emph{masks}, one-to-one corresponding to the decomposition ranks, $\bm{G} \odot \bm{m} = \left\{\mathcal{G}_k \odot \bm{m} \right\}$ is a set of \emph{masked cores}:
\begin{equation*}
    \mathcal{G}_k \odot \bm{m} \coloneqq \mathcal{G}_k \odot_{k_1} m_{k_1} \dots \odot_{k_p} m_{k_p},
\end{equation*}
$r_{k_1},\dots,r_{k_p} \in \dims\left(\mathcal{G}_k\right)$ are all ranks belonging to the dimensions of the core $\mathcal{G}_k$. 
The likelihood $p\left(y \mid x, \bm{G} \odot \bm{m}\right)$ is defined by the tensorized model, so, for example, it can be (exponent of negative) cross-entropy loss for classification tasks. 
Note that in model~\eqref{eq:model}, we completely remove the dependency between cores and masks unlike, e.g., the widely known spike-and-slab prior model~\citep{ishwaran2005spike}.
This allows to ensure universality of our approach by accepting arbitrary couplings of the cores in general decompositions. 

We assume the factorized Bernoulli prior over masks with the success probability $\pi$, which is a natural hyperparameter of our model regulating the intensity of compression: 
\begin{equation}
    \label{eq:prior}
    p\left(\bm{m} \right) = p\left(\bm{m} \mid \pi\right) = \prod_{k} \prod_{s=1}^{r_k} \pi^{m_k(s)} (1-\pi)^{1 - m_k(s)}.
\end{equation}
This prior term is the key ingredient that allows achieving low-rank solutions as it sparsifies the selection masks. 
Note that instead of adjusting several or even dozens of ranks in the decomposition, one needs to validate only one hyperparameter in our model (along with careful initialization). 
Furthermore, in our experiments, we found that $\pi$ does influence the final compression-accuracy trade-off but not crucially (see~\Cref{app:pi_ablation}).
Taking $\pi \approx 10^{-2}$ is usually a good choice. 
Note that setting $\pi$ to values more than $0.5$ is of no use, as therefore, the prior term would foster dense masks, which contradicts with obtaining a low-rank solution corresponding to mostly zero-valued masks.

We also put the factorized zero-mean Gaussian with a large variance as the prior distribution over the cores $p\left(\bm{G}\right)$.
It serves as a slight $L_2$ regularization and we have empirically found that it helps to balance the coefficients in the cores, stabilize the training process and improve test accuracy.
We did not conduct a thorough search for the optimal values of the prior variance and used the same fixed value of $10^2$ in all our experiments.

In this work, we consider finding \emph{maximum a posteriori} (MAP) estimates of parameters $\bm{G}$ and $\bm{m}$ in model~\eqref{eq:model}:
\begin{multline}
\label{eq:map}
\sum_{i=1}^N \log p\left(y_i \mid x_i, \bm{G} \odot \bm{m}\right) + \log p\left(\bm{m}\right) +\\
+ \log p\left(\bm{G}\right) \longrightarrow \max_{\bm{m}, \bm{G}}.
\end{multline}
Naturally, this problem implies discrete optimization over binary masks and, hence, is infeasible due to exhaustive search in the general case. 
To tackle this, we first substitute problem~\eqref{eq:map} with an equivalent:
\begin{multline}
\label{eq:elbo_map}
\E_{\bm{m} \sim q(\bm{m})} \left[ \sum_{i=1}^N \log p\left(y_i \mid x_i, \bm{G} \odot \bm{m}\right) + \log p\left(\bm{m}\right) \right] + \\
+ \log p\left(\bm{G}\right) \longrightarrow \max_{q(\bm{m}), \bm{G}},
\end{multline}
where the family of distributions $q(\bm{m})$ includes all deterministic ones, i.e., taking only a single value. 
The solutions of problems~\eqref{eq:map} and~\eqref{eq:elbo_map} coincide according to \Cref{lemma:exp_max}. 

\begin{lemma}
\label{lemma:exp_max}
For an arbitrary scalar function $F(x)$ with attainable maximum the following problems are equivalent:
\begin{equation*}
    \max_x F(x) \equiv \max_{q(x)} \E_{x \sim q(x)} \left[ F(x) \right]
\end{equation*}
if the family of distributions $q(x)$ includes degenerate ones.
\end{lemma}
\begin{proof}
The statement follows from the fact that for \emph{any} distribution $q(x)$ we have
\begin{equation}
    \label{eq:exp_max}
    \E_{x\sim q(x)} F(x) \le F(x^*),
\end{equation}
where $x^* = \arg\!\max_x F(x)$, and \cref{eq:exp_max} turns into equality when $q(x) = \delta(x - x^*)$.
\end{proof}

Next, we constrain $q(\bm{m})$ to be a factorized Bernoulli distribution over each mask element $m_k(s)$ with parameters $\bm{\phi} = \{\phi_k(s)\}$. 
Note that this family meets the requirement to include all degenerate solutions in order to ensure equivalence of \cref{eq:map} and \cref{eq:elbo_map}.
Now the problem~\eqref{eq:elbo_map} translates into the following:
\begin{multline}
    \label{eq:elbo_bern_map}
    \E_{\bm{m} \sim q_{\bm{\phi}}(\bm{m})} \left[ \sum_{i=1}^N \log p\left(y_i \mid x_i, \bm{G} \odot \bm{m}\right) \right] + \\
    + \sum_k \sum_{s = 1}^{r_k} \left[ \phi_k(s) \log \pi + (1 - \phi_k(s)) \log (1 - \pi)\right] + \\
    + \log p\left(\bm{G}\right) \longrightarrow \max_{\bm{\phi}, \bm{G}}.
\end{multline}
One can notice that adding the $q$ entropy term into \cref{eq:elbo_bern_map} yields the evidence lower bound (ELBO) maximization, a well-known Bayesian technique for the variational posterior approximation, with a factorized Bernoulli variational distribution.
However, we \emph{do not} perform variational inference with MARS, but look for a single-point solution instead.
We discuss this further at the end of the article.

We solve the maximization problem~\eqref{eq:elbo_bern_map} with the stochastic gradient method. 
To calculate low-variance stochastic gradients w.r.t. parameters $\bm{\phi}$ in \cref{eq:elbo_bern_map}, we use the \emph{reparameterization trick}~\citep{kingma2013auto}. 
To this end, we soften the discrete samples from $q_{\bm{\phi}}(\bm{m})$ in the expectation term via the Binary Concrete relaxation~\citep{maddison2017concrete, jang2017categorical} with temperature, which defines ``discreteness'' of the relaxed samples, decaying to zero in the course of training.
After training, we round the probabilities $\bm{\phi}$ to binary masks $\bm{m}_{MAP}$ and use a compact solution with reduced cores $\bm{G}_{MAP} \odot \bm{m}_{MAP}$ to predict for a new data sample $x^*$: $p\left(y^* \mid x^*, \bm{G}_{MAP} \odot \bm{m}_{MAP}\right)$.

\Cref{alg:mars_opt} summarizes the training procedure.
$RB(\phi, \tau)$ denotes the Relaxed Bernoulli distribution, which is Binary Concrete with temperature $\tau$ and location $\frac{\phi}{1 - \phi}$.
Sampling from $RB(\phi, \tau)$ is as simple as applying a differentiable operation 
\[\sigma \left(\frac{\log(u) - \log(1 - u) + \log(\phi) - \log(1 - \phi)}{\tau} \right)\]
to $u \sim \mathcal{U}_{[0,1]}$, where $\sigma(x) = 1 / (1 + e^{-x})$ is a logistic sigmoid function.
Note that~\Cref{alg:mars_opt} is essentially an SGD on a regularized loss, therefore they have similar computational complexity; it can be considered extremely light-weight compared to other (e.g., SVD-based) rank selection schemes.

\begin{algorithm}
   \caption{MARS relaxed MAP learning procedure}
   \label{alg:mars_opt}
\begin{algorithmic}
   \STATE {\bfseries Input:} data $\left(X, Y\right)$, prior parameter $\pi$, temperature $\tau$, batch size $B$
   
   \STATE {\bfseries Output:} MAP estimate of cores $\bm{G}_{MAP}$ and masks $\bm{m}_{MAP}$
   
   \STATE Initialize $\bm{G}$ and $\bm{\phi}$
   
   \REPEAT
   \STATE Sample masks $\hat{\bm{m}} = \{\hat{m}_k(s) \sim RB(\phi_k(s), \tau)\}$
   
   \STATE Sample a mini-batch of objects $\{(x_{i_l}, y_{i_l})\}_{l=1}^B$
   
   \STATE $L \coloneqq \sum_{l=1}^B \log p\left(y_{i_l} \mid x_{i_l}, \bm{G} \odot \hat{\bm{m}}\right)$
   
   \STATE $g_{\bm{\phi}} \coloneqq \frac{\partial L}{\partial \bm{G} \odot \hat{\bm{m}}} \frac {\partial \bm{G} \odot \hat{\bm{m}}}{\partial \hat{\bm{m}}} \frac{\partial \hat{\bm{m}}}{\partial \bm{\phi}} + \log \left( \frac{\pi}{1- \pi} \right)$
   
   \STATE $g_{\bm{G}} \coloneqq \frac{\partial L}{\partial \bm{G} \odot \hat{\bm{m}}} \frac {\partial \bm{G} \odot \hat{\bm{m}}}{\partial \bm{G}} + \frac{\partial \log p \left(\bm{G}\right)}{\partial \bm{G}}$
   
   \STATE Update $\bm{\phi}$ using stochastic gradient $g_{\bm{\phi}}$
   
   \STATE Update $\bm{G}$ using stochastic gradient $g_{\bm{G}}$
   
   \STATE Decay $\tau$
   \UNTIL{stop criterion is met}
   \STATE Define $\bm{G}_{MAP} \coloneqq \bm{G}$
   \STATE Define $\bm{m}_{MAP} \coloneqq round(\bm{\phi})$
\end{algorithmic}
\end{algorithm}

\section{EXPERIMENTS}

In this section, we present the results of the conducted experiments with MARS that show its ability to improve previous results on DNN tensorization.
In \Cref{app:tucker_approx}, we also provide additional experiments involving a different tensorized model to demonstrate that MARS is a general ranks selection scheme not confined to TNNs only.

We train tensorized models using MARS according to \Cref{alg:mars_opt}.
The learned hard binary masks are then applied to the trained cores to remove excess ranks and obtain a compact architecture that is used during test-time inference.
Careful parameter initialization is required for optimal performance and training.
We propose to initialize logits of $\bm{\phi}$ using the normal distribution centered at some value $\alpha$, which is an important hyperparameter with a role similar to that of hyperparameter $\pi$.
We refer to \Cref{app:imp_details} for more details on implementation.

Our experiments are conducted in three ways to prove the efficiency and versatility of MARS. 
First, we show how MARS can restore the actual ranks when the ground truth (GT) is known (\Cref{sec:toy_exp}, \Cref{app:tucker_approx}).
Second, we compare with alternative approaches for Bayesian rank selection (\Cref{sec:mnist_2fc,sec:mnist_lenet,sec:cifar_resnet}) to show that MARS can perform just as well, being a more general method. 
Third, we apply MARS to various practical tensorized models to show that it can significantly improve previous results where ranks were mostly selected using cross-validation (\Cref{sec:sent_anal,sec:mnist_lenet}).
In tables, we report mean $\pm$ std where applicable.

\subsection{Toy experiment}
\label{sec:toy_exp}

Our first experiment serves as a mere proof of concept.
We evaluate our method on a toy linear classification task with a factorized parameter matrix to verify how MARS can approximate the true rank.

Let $N$, $D$, $C$, $r^*$, $R$ denote, respectively, the number of samples, input, output dimensions, the ground truth rank of the problem, and the initial rank of the parameters to be reduced.
At first, we sample elements of the input matrix $X \in \R^{N \times D}$ i.i.d. from the standard normal distribution.
We similarly sample the ground truth parameter matrices $U^* \in \R^{D \times r^*}$ and $V^* \in \R^{r^* \times C}$.
After that, we obtain the output matrix $Y = \OHE \left(XU^*V^*\right)$ of size $(N, C)$, where $\OHE$ denotes row-wise argmax one-hot encoding operation.
Finally, we initialize the learnable parameter matrices $U \in \R^{D \times R}$ and $V \in \R^{R \times C}$ and train a linear classifier with a factorized parameter matrix $UV$ using MARS to reduce the initial rank $R > r^*$ and restore the GT rank $r^*$.

We fixed $N=10000$, $D=128$, $C=32$, $R=32$ and varied $r^*$.
Namely, we considered three cases: $r^* = 8$, $r^* = 12$, and $r^* = 16$.
We took $\pi = 10^{-2}$ and $\alpha = -4$, $\alpha = -3.5$, and $\alpha = -3$ for each case, respectively.
We evaluated models on a separate test set and trained a vanilla linear classifier as a baseline.\footnote{We have also tried a factorized baseline with $R = 32$ but obtained much worse results and decided not to include it.}

We report the results averaged over 10 runs in \Cref{tab:toy}.
As can be seen, MARS can rather accurately restore the true rank $r^*$ starting from a higher initial value $R$. 
Furthermore, the obtained models are not only more compact but also exhibit better test accuracy than the baseline.

As neural network training encounters a manifold of different local minima, the problem of revealing the ``true rank'' of a tensorized DNN, rather than a simple linear model, is ill-posed: it highly depends on initialization, optimization, and hyperparameters. 
Yet, that can be especially useful for \emph{ensembling}, which will be discussed further.

\begin{table}
\caption{True rank restoring with MARS in a toy linear classification task with a factorized parameter matrix.
Results are averaged over 10 runs.} 
\label{tab:toy}
\centering
\begin{tabular}{cccc}
\toprule
GT rank $r^*$ & Reduced $R$ & Accuracy & Baseline \\
\midrule
$8$ & $8.4 \pm 0.5$ & \bm{$91.8 \pm 0.6 \%$} & $87.3\%$ \\
$12$ & $12.6 \pm 0.7$ & \bm{$89.5 \pm 0.7\%$} & $85\%$ \\
$16$ & $18 \pm 1.3$ & \bm{$85.4 \pm 0.7\%$} & $82.8\%$ \\
\bottomrule
\end{tabular}
\end{table}

\subsection{MNIST 2FC-Net}
\label{sec:mnist_2fc}

In this experiment, we compare MARS against the LR-BTNN method of~\citet{hawkins2021bayesian} on the MNIST~\citep{lecun1998mnist} dataset.
In this task, both methods aim to automatically select ranks in a relatively small image classification neural network with two TT-decomposed fully-connected layers of sizes $784 \times 625$ and $625 \times 10$.
As proposed by~\citet{hawkins2021bayesian}, we take the following dimensions factorization of the TT-layers: $(n_1, n_2, n_3, n_4) = (7, 4, 7, 4)$, $(m_1, m_2, m_3, m_4) = (5, 5, 5, 5)$ and $(n_1, n_2) = (25, 25)$, $(m_1, m_2) = (5, 2)$ for the first and second layer, respectively. 
All the initial ranks are set to $20$, which gives $18\times$ compression at the start.

\begin{table}[ht]
\caption{Compression/accuracy on MNIST with 2FC-Net.
Results are averaged over 10 runs.} 
\label{tab:mnist_2fc}
\centering
\begin{tabular}{lcc}
\toprule
Model & Compression & Accuracy \\
\midrule
Baseline & $1\times$ & $\bm{98.2\%}$ \\
Baseline-TT & $18\times$ & $97.7\%$ \\
LR-BTNN & $137\times$ & $97.8\%$ \\
MARS (soft) & $141 \pm 18.6 \times$ & $\bm{98.2 \pm 0.11\%}$ \\
MARS (hard) & $\bm{205 \pm 30.9 \times}$ & $97.9 \pm 0.19\%$ \\
\bottomrule
\end{tabular}
\end{table}

We execute MARS in two modes for this task: soft compression mode ($\alpha=-1.5$, $\pi=10^{-1}$) and hard compression mode ($\alpha=-1.75$, $\pi=10^{-2}$).
In each mode, we trained 10 networks from different random initializations and averaged the results.
\Cref{tab:mnist_2fc} shows that MARS surpasses the approach of~\citet{hawkins2021bayesian} in this task both in terms of compression and final accuracy, even though LR-BTNN is specifically tailored for the Tensor Train decomposition. 

We would also like to note that an ensemble of small MAP networks, obtained in the soft compression mode, gives the accuracy of $\bm{98.9\%}$. 
We argue that compact TNN ensembling might be a promising research direction.

\begin{figure}[ht]
\centering
\includegraphics[width=1.0\linewidth]{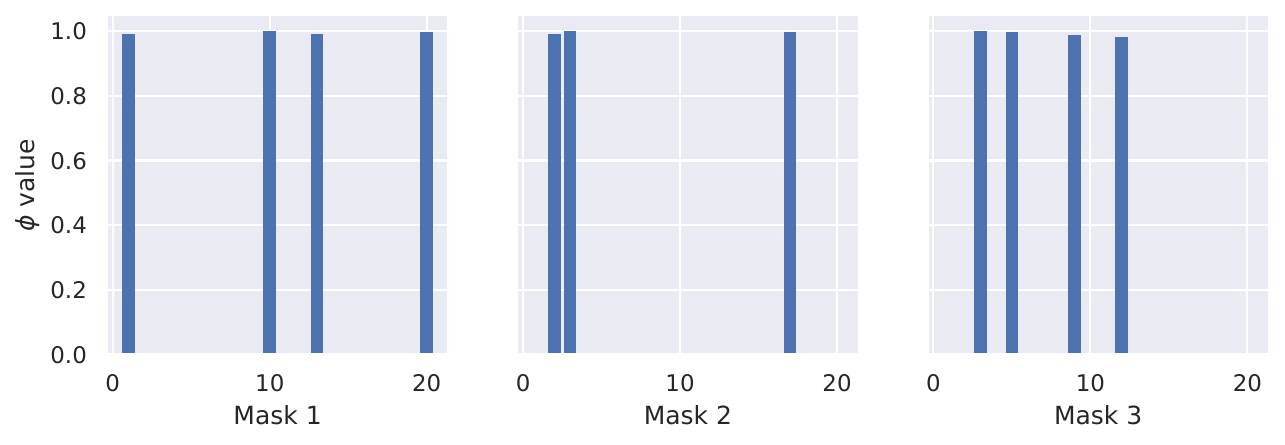}
\caption{Learned binary masks probabilities $\bm{\phi}$ corresponding to the first TT-layer in MNIST 2FC-Net. 
Note that the relaxed MARS MAP estimate is quite close to the deterministic solution.}
\label{fig:mnist_ttfc1}
\end{figure}

\Cref{fig:mnist_ttfc1} shows the bar plot of $\bm{\phi}$ values of the three masks corresponding to the first TT-layer.
We see that the relaxed MAP estimate is actually quite close to the deterministic binary masks. 
After rounding to strictly binary values and applying the resulted masks to the TT-cores, the ranks of the first layer shrink to $(r_0, r_1, r_2, r_3, r_4) = (1, 4, 3, 4, 1)$ which leads to more than $556\times$ layer compression.

\subsection{Sentiment analysis with TT-embeddings}
\label{sec:sent_anal}

A recent work of~\citet{hrinchuk2020tensorized} leverage Tensor Train decomposition for compressing embedding layers in various NLP models. 
The authors propose to convert the matrix of embeddings into the TT-format alike TT-FC layers.
They provide a heuristic to automatically determine optimal (in terms of the occupied memory) factorization of dimensions in TT-matrices given the number of factors $d$.
However, in their experiments, the ranks in the TT-decomposition were still manually set equal to some predefined value.

We repeat their experiment on the sentiment analysis task and apply MARS on top of the tensorized model.
The model consists of a TT-embedding layer with ranks equal to $16$, followed by an LSTM, which performs sentiment classification. 
The authors evaluated on two datasets: IMDB~\citep{maas2011learning} and Stanford Sentiment
Treebank (SST)~\citep{socher2013recursive}. 
On each dataset they tried three tensorized models: with $d=3$, $d=4$ and $d=6$ factors in the TT-matrix of embeddings, respectively.
On IMDB, the authors obtained both maximal accuracy and compression with the model using $d=6$ factors.
On SST, the best compression was achieved at $d=6$, while the best accuracy was achieved at $d=3$.
Thus, we choose the best model on IMDB and the medium one ($d = 4$) on SST and train them with MARS.
We set $\pi=10^{-2}$ in both models and $\alpha=-0.25$, $\alpha=-1.0$ for the first and the second one, respectively.

\Cref{tab:sent} presents the obtained results. 
Automatic ranks selection with MARS allowed to significantly improve both quality and compression of the best IMDB TT-model.
On SST, we managed to overtake the best compressing and best performing models with a medium model trained using our method.
The final selected ranks are $(r_0, r_1, r_2, r_3, r_4, r_5, r_6) = (1, 8, 11, 15, 16, 16, 1)$ and $(r_0, r_1, r_2, r_3, r_4) = (1, 6, 14, 14, 1)$ for IMDB and SST MARS TT-models, respectively.
We hypothesize that such an escalating rank distribution could be explained by the hierarchical indexing in TT-embeddings, where first TT-cores are responsible for indexing large blocks in the embedding matrix, and subsequent cores index inside those blocks. 
The compressed model might find only a few large blocks in the whole embedding matrix relevant for prediction, thus, the first cores could be made less expressive.
On the whole, one can see that setting decomposition ranks equal, which is a common heuristics in tensorized networks, is quite inefficient as opposed to the nonuniform ranks selection.

\begin{table}[ht]
\caption{Compression and accuracy on sentiment analysis with TT-embedding layers. 
TT-$d$ denotes TT-embedding with $d$ factors.}
\label{tab:sent}
\centering
\begin{tabular}{@{}llcc@{}}
\toprule
Dataset & Model & Compression & Accuracy \\ \midrule
\multirow{3}{*}{IMDB} & Baseline & $1\times$ & $88.6\%$ \\
 & TT-$6$ & $441\times$ & $89.7\%$ \\
 & MARS + TT-$6$ & $\bm{559\times}$ & $\bm{90.1\%}$ \\ \midrule
\multirow{4}{*}{SST} & Baseline & $1\times$ & $37.4\%$ \\
 & TT-$3$ & $78\times$ & $41.5\%$ \\
 & TT-$6$ & $307\times$ & $39.9\%$ \\
 & MARS + TT-$4$ & $\bm{340\times}$ & $\bm{42.4\%}$ \\ \bottomrule
\end{tabular}
\end{table}

\subsection{MNIST LeNet-5}
\label{sec:mnist_lenet}

\begin{table*}[ht]
\caption{Compression, accuracy, and speed-up on MNIST with LeNet-5.
TRN-$r$ denotes the TRN model with the same Tensor Ring rank $r$.
Speed-up is evaluated as the ratio of test time per 10000 samples of the baseline and the given model, as proposed in~\citet{wang2018wide}.
Results are averaged over 5 runs.} 
\label{tab:mnist_lenet5}
\centering
\begin{tabular}{lccc}
\toprule
Model & Compression & Accuracy & Speed-up \\
\midrule
Baseline & $1\times$ & $99.2\%$ & $1.0\times$ \\
Tucker & $2\times$ & $99.1\%$ & $0.58\times$ \\
TRN-$10$ & $\bm{39\times}$ & $98.6\%$ & $0.48\times$ \\
TRN-$15$ & $18\times$ & $99.2\%$ & $0.97\times$ \\
TRN-$20$ & $11\times$ & $\bm{99.3\%}$ & $0.73\times$ \\
MARS + Tucker & $10 \pm 0.8\times$ & $99.0 \pm 0.07\%$ & $\bm{1.19 \pm 0.01\times}$ \\
\bottomrule
\end{tabular}
\end{table*}

In~\citet{wang2018wide} Tensor Ring (TR) decomposition~\citep{zhao2016tensor}, a generalization of the Tensor Train decomposition, was applied to compress convolutional networks.
Such neural networks with TR-decomposed convolutions and FC-layers are called Tensor Ring Nets (TRNs).
The authors compared their approach against~\citet{DBLP:journals/corr/KimPYCYS15}, where Tucker-2 and low-rank matrix factorization, which represent a simpler decomposition family, are used for the same purpose.
In one of the experiments, both methods were evaluated on compressing and accelerating LeNet-5~\citep{lecun1998gradient}, a relatively small convolutional neural network with 2 convolutional layers, followed by 2 fully-connected layers, on the MNIST dataset. 
TRN significantly surpassed the simpler Tucker approach.

In this experiment, we demonstrate that even using less expressive types of decompositions, one can achieve results comparable to TRN by training with MARS. 
Namely, we apply Tucker-2 decomposition to the second convolution and low-rank factorization to the first FC-layer, as the other layers occupy less than $1.3\%$ of all model parameters.
We automatically select the two Tucker-2 decomposition ranks $r_1$, $r_2$ and the matrix rank $r_3$ using our method, starting from $r_1= r_2 = 20$, $r_3 = 100$ ($2.9\times$ compression at the start).
We initialize the mean value of $\bm{\phi}$ logits $\alpha$ with zero and set $\pi = 10^{-2}$. 

The averaged results over 5 runs are presented in \Cref{tab:mnist_lenet5}.
MARS enhanced compression of the Tucker model by a factor of $5$ with about the same quality, making it comparable to TRN, which is based on a significantly more complex decomposition family.
We would like to note that the Tucker model already has an inner mechanism of ranks selection, yet, it can only approximate the ranks required for the layers decomposition, after which the model is fine-tuned.
MARS performs end-to-end ranks selection with model training, which leads to significantly better results.

Another important achievement of our model is the ability to actually accelerate networks.\footnote{We do not specify run times in other experiments, as this information is not provided in the related works; however, we have observed acceleration in other cases as well.} 
Even though TR-decomposition allows achieving better compression, it, however, slows down the inference.
The authors argue that such an effect is caused by the suboptimality of the existing hard- and software for tensor routines.
Using simpler layer factorizations, we managed to speed up LeNet-5 by $1.2\times$.

Similarly to the previous experiment, we measured the quality of the ensemble of LeNet-5 networks compressed with MARS.
Ensembling aids to improve model test accuracy up to $\bm{99.5\%}$.
Note that the ensemble of 5 networks compressed by $10\times$ still requires twice less memory than the original model and, provided parallel computing, can even work faster.

We recognize the power of the Tensor Ring decomposition in compressing neural networks.
Since in TRN all decomposition ranks are set equally, we believe that MARS could further improve its results and leave it for future work.

\subsection{CIFAR-10 ResNet-110}
\label{sec:cifar_resnet}

The main experiment of~\citet{hawkins2021bayesian} consisted in applying their LR-BTNN method to ResNet-110~\citep{DBLP:conf/cvpr/HeZRS16} on CIFAR-10 dataset~\citep{krizhevsky2014cifar}.
The authors used Tensor Train decomposition for compressing all convolutional layers except for the first ResNet block (first 36 layers) and the $1\times1$ convolutions.

However, they implemented a simplified scheme of decomposing convolutions, which we call \emph{naive}.
At first, the numbers of input and output channels $N$ and $M$ are factored into $N = \prod_{k=1}^d n_k$, $M = \prod_{k=1}^d m_k$. 
After that, the 4-dimensional convolutional kernel with kernel size $k$ is reshaped into a $(2d+1)$-way tensor with dimensions $(n_1, \dots, n_d, m_1, \dots, m_d, k^2)$.
The reshaped tensor is then decomposed into Tensor Train with $2d+1$ cores.
Such a scheme could be fruitful in terms of compression, yet it does not have a potential for efficient computing due to the need to construct the full convolutional tensor from the TT-cores on each forward pass.
Unlike this method, \citet{garipov16ttconv} proposed to represent convolutions as $k^2 N \times M$ matrices in TT-format based on the fact that most frameworks reduce the convolution operation to a matrix-by-matrix multiplication.
We call the scheme of~\citet{garipov16ttconv} \emph{proper}.
This approach, for instance, was leveraged to achieve more than $4\times$ better energy efficiency and $5\times$ acceleration compared to state-of-the-art solutions on a special TT-optimized hardware~\citep{deng2019tie}.

We repeat the ResNet experiment of~\citet{hawkins2021bayesian} with MARS using both naive and proper schemes for TT-decomposition of convolutions.
The original paper does not provide much detail on the experiment setting, however, we could deduce that the authors used $d=2$ and $d=3$ factors for the second and third ResNet block, respectively, i.e., in the second block, they reshaped convolutional kernels from $(32, 32, 3, 3)$ to $(8, 4, 8, 4, 9)$ and in the third one from $(64, 64, 3, 3)$ to $(4, 4, 4, 4, 4, 4, 9)$.
In order to obtain similar number of TT-cores for the proper scheme, we choose the following respective shapes of convolutional TT-matrices: $(2, 2) \times (3, 2) \times (3, 2) \times (4, 2) \times (4, 2)$ and $(2, 2) \times (2, 2) \times (3, 2) \times (3, 2) \times (4, 2) \times (4, 2)$.
At the start, all ranks equal $20$, which gives $2.7\times$ and $2.3\times$ compression of the naive and proper models, respectively.
We set $\pi = 10^{-2}$, $\alpha=2.25$ and $\pi = 4 \cdot 10^{-3}$, $\alpha=3.0$ in those models, respectively.

The results are given in \Cref{tab:cifar_resnet}. 
Using the naive scheme, MARS achieved the results comparable to LR-BTNN: it performed slightly worse in compression but better in accuracy.
Proper TT-decomposition of convolutions and training with MARS allowed to reach the same quality as with the naively decomposed baseline TT-model but at a significantly higher compression ratio, which once again emphasizes the efficiency of the~\citet{garipov16ttconv} scheme and nonuniform rank distribution in tensorized models.

\begin{table}[ht]
\caption{Compression and accuracy on CIFAR-10 with ResNet-110. 
We put the type of the used decomposition scheme in parentheses.}
\label{tab:cifar_resnet}
\centering
\begin{tabular}{lcc}
\toprule
Model & Compression & Accuracy \\
\midrule
Baseline & $1\times$ & $\bm{92.6\%}$ \\
Baseline (naive) & $2.7\times$ & $91.1\%$ \\
LR-BTNN (naive) & $\bm{7.4\times}$ & $90.4\%$ \\
MARS (naive) & $7.0\times$ & $90.7\%$ \\
MARS (proper) & $5.5\times$ & $91.1\%$ \\
\bottomrule
\end{tabular}
\end{table}

\section{CONCLUSION AND DISCUSSION}

In this paper, we present MARS, the method for efficient automatic selection of ranks in tensorized models leveraging arbitrary tensor decompositions. 
The basic principle of MARS is learning binary masks along with overall model training that cover the cores of the decomposition and automatically select the optimal structure.
We perform learning of masks and model parameters via relaxed MAP estimation in a special Bayesian probabilistic model.
The experiments demonstrate that our technique can improve the accuracy and compression of tensorized models with manually selected ranks and surpasses or performs comparably with alternative rank selection methods specialized on concrete types of tensor decompositions. 

It is widely known that the ensembling of deep neural networks leads to significant quality improvement~\citep{lakshminarayanan2017simple, lobacheva2020power}.
In our experiments, we observed a similar trend with ensembles of compact MARS-trained networks. 
However, usual DNN ensembles require training and evaluating several neural networks, which might be inapplicable in resource-constrained environments.
By contrast, the whole ensemble of tensorized networks often occupies less memory than a single standard model.
This opens a very promising prospect for future research.

MARS obtains a single MAP estimate of masks.
However, learning the variational distribution over binary masks or sampling from the posterior could allow efficient ensembling of compact tensorized models.
We noted in \Cref{sec:method} that our objective~\eqref{eq:elbo_bern_map} resembles ELBO up to the entropy term.
Unfortunately, our preliminary experiments in variational inference with factorized Bernoulli $q_{\bm{\phi}}(\bm{m})$ led to distributions with overly low variance. 
In other words, sampling from $q_{\bm{\phi}}(\bm{m})$ did not improve accuracy compared to the model spawned by its mode.
This might mean that fully-factorized Bernoulli distribution cannot appropriately approximate the true posterior due to numerous correlations between mask variables. 
However, it is quite effective for finding the MAP estimate. 
We believe that more flexible variational families, e.g., those based on hierarchical models, may better approximate the posterior, and leave it for future study.

Other research directions may include improvements of the model and the learning method, for instance, trying REINFORCE-like algorithms~\citep{williams1992simple} for optimization over discrete masks.
Applying MARS to other types of tensor decompositions (like TR-decomposition) and tensorized models is also very intriguing.

In~\Cref{app:limitations}, we provide more discussion on the limitations of this research, its perspectives, and possible societal impact.

\subsubsection*{Acknowledgements}

We would like to thank Artem Grachev for the valuable discussions on the original idea of selecting ranks in tensor decompositions.
We are also very grateful to the reviewers for their constructive and helpful comments on the submitted manuscript.
The publication was supported by the grant for research centers in the field of AI provided by the Analytical Center for the Government of the Russian Federation (ACRF) in accordance with the agreement on the provision of subsidies (identifier of the agreement 000000D730321P5Q0002) and the agreement with HSE University \textnumero 70-2021-00139.

\bibliographystyle{apalike}
\bibliography{ourbib}

\appendix
\onecolumn

\section{ADDITIONAL RELATED WORK}
\label{app:rel_work}

In this section, we provide additional related work on neural networks tensorization and optimal decomposition topology estimation, which goes beyond the scope of our study.

Tensor methods allow achieving significant compression, acceleration, and sometimes even quality improvement of neural networks.
In~\citet{DBLP:journals/corr/LebedevGROL14}, 4-dimensional convolutional kernel tensors are decomposed with CP decomposition. 
The authors were able to accelerate a network by more than 8 times without significantly decreasing accuracy.
In~\citet{novikov15tensornet}, TT-decomposition was leveraged to achieve up to $200000\times$ compression of fully-connected layers in a VGG-like network.
\citet{hrinchuk2020tensorized} used a similar approach to compress embedding layers in NLP models, which in some cases led to a noticeable quality increase due to the induced regularization. 
In~\citet{DBLP:conf/icml/YangKT17} the authors managed to achieve comparable performance with state-of-the-art models on very high-dimensional video classification tasks using orders of magnitude less complex TT-tensorized recurrent neural networks.
Recently,~\citet{ma2019tensorized} applied Block-Term tensor decomposition (BTD)~\citep{de2008decompositions}, a combination of CP and Tucker decompositions, to efficiently compress Multi-linear attention layers in Transformers and improved the single-model SoTA in language modeling.
However, in all of these works, ranks selection was done manually for each decomposed layer.

A series of works has recently emerged based on greedy/evolutionary algorithms to learn the optimal tensor network topology~\citep{NEURIPS2019_2bd2e337, li2020evolutionary, hashemizadeh2020adaptive, li2021rank, li2022permutation}. 
These methods tackle a more general problem than optimal ranks selection in a concrete decomposition and, hence, impose overly complex multi-step algorithms to be directly applied for full DNN tensorization (mainly, these approaches consider simpler tasks like plain tensor decomposition or tensor completion).

\section{TENSORIZED LAYERS}
\label{app:tens_layers}

In this section, we provide details concerning different tensorized neural network layers used in this work. 

The simplest example of a decomposed layer is a fully-connected layer approximated via low-rank matrix factorization.
In this case the matrix of weights $W \in \mathbb{R}^{M \times N}$ is represented via contraction (or matrix product) of two low-rank matrices $U_1 \in \mathbb{R}^{M\times r}$ and $U_2 \in \mathbb{R}^{r \times N}$:
\begin{equation*}
    W = U_1  U_2.
\end{equation*}
Mapping the input $x \in \mathbb{R}^N$ through these matrices in series leads to FLOPs reduction from $O\left(MN\right)$ to $O\left( r (M + N) \right)$, which could give a significant gain when $r$ is smaller than $M$ and $N$.

Similarly, Tucker-2 decomposition of a convolutional kernel~\citep{DBLP:journals/corr/KimPYCYS15} results in three consecutive smaller-sized convolutions. 
Namely, the convolutional kernel $\mathcal{K} \in \mathbb{R}^{C_{in} \times C_{out} \times k \times k}$, where $C_{in}$, $C_{out}$ are the numbers of input and output channels and $k$ is the kernel size, decomposes into two matrices $U_1 \in \mathbb{R}^{C_{in} \times r_1}$, $U_2 \in \mathbb{R}^{C_{out} \times r_2}$ and a smaller 4-dimensional tensor $\mathcal{G} \in \mathbb{R}^{r_1 \times r_2 \times k \times k}$ via the partial Tucker decomposition as:
\begin{equation*}
    \mathcal{K} = \mathcal{G} \times_1 U_1 \times_2 U_2.
\end{equation*}
Convolution operation with such a kernel can be rewritten as the following series of simpler convolutions: $1 \times 1$-convolution, reducing the number of channels from $C_{in}$ to $r_1$, $k \times k$-convolution with $r_1$ input and $r_2$ output channels and again $1 \times 1$-convolution, restoring the number of output channels from $r_2$ to $C_{out}$.
This trick helps to compress and speed up convolutions when the number of intermediate channels (i.e., ranks) is smaller than $C_{in}$ and $C_{out}$.

In a fully-connected TT-layer (TT-FC)~\citep{novikov15tensornet}, the matrix of weights $W \in \mathbb{R}^{M \times N}$, input and output vectors $x \in \mathbb{R}^N$ and $y \in \mathbb{R}^M$ are reshaped into tensors $\mathcal{W} \in \mathbb{R}^{(m_1, n_1) \times \dots \times (m_d, n_d)}$, $\mathcal{X} \in \mathbb{R}^{n_1 \times \dots \times n_d}$ and $\mathcal{Y} \in \mathbb{R}^{m_1 \times \dots \times m_d}$, respectively, where $M = \prod_{k=1}^d m_k$, $N = \prod_{k=1}^d n_k$. 
Then $\mathcal{W}$ is converted into the TT-format with 4-dimensional cores $\bm{G} = \left\{ \mathcal{G}_1, \dots, \mathcal{G}_d  \right\}$, $\mathcal{G}_k \in \mathbb{R}^{r_{k-1} \times m_k \times n_k \times r_k}$.
The linear mapping $y = W x$ translates into a series of contractions:\footnote{
Strictly speaking, contractions over two modes $n_k$ and $r_k$.}
\begin{equation*}
    \mathcal{Y} = \mathcal{G}_1 \dots \mathcal{G}_d \mathcal{X},
\end{equation*}
which, calculated from right to left, yields the computational complexity $O\left(dr^2n\max\{M, N\}\right)$, where $r$ is the maximal TT-rank, $n = \max_{k = 1 \dots d} n_k$. 
Similar technique, based on matrices represented in TT-format, or \emph{TT-matrices}, underlies most other types of TT-layers.

\section{IMPLEMENTATION DETAILS}
\label{app:imp_details}

Our implementation is based on \texttt{tt-pytorch}\footnote{\url{https://github.com/KhrulkovV/tt-pytorch}} library~\citep{khrulkov2019tensorized}, which provides the minimal required tools for working with TT-decomposition in neural networks using \texttt{PyTorch}~\citep{paszke2019pytorch}. 

\paragraph{Initialization} 
We use the Glorot-like~\citep{DBLP:journals/jmlr/GlorotB10} initialization for the TT-cores, implemented in the library and described in the corresponding paper, and the Kaiming Uniform initialization~\citep{he2015delving} for the Tucker-2 cores and matrices, which is default in \texttt{PyTorch}. 
We discovered that initialization and parameterization of masks probabilities matter: we use the logit reparameterization and initialize logits of $\bm{\phi}$ from the normal distribution with scale $10^{-2}$ and mean $\alpha$, which is a hyperparameter (concrete values were chosen with a simple cross-validation procedure and are provided in each experiment).
Variance of the normal prior over cores $p \left( \bm{G} \right)$ is fixed and equals $10^2$.

Choosing the initial rank is generally a non-trivial problem that can be attributed to common sparsifying/pruning techniques. 
We can state, however, that MARS does not significantly depend on this hyperparameter (as long as it is set somewhat reasonably) and is able to restore the ground truth rank from different initializations, as, e.g., demonstrated in~\Cref{sec:toy_exp} in the main text. 
We have empirically observed that running MARS from a higher initial rank with a slight tuning of $\pi$ and/or $\alpha$ hyperparameters produces similar results in most cases.

\paragraph{Training}

In practice, to assist optimization, we do not multiply each of the decomposition cores, coupled via a shared mode, by the same corresponding relaxed binary mask, but instead perform only one multiplication. 
For instance, in Tucker-2 convolutional layer with masks $\bm{m} = \{m_1, m_2\}$ we apply the respective mask multiplication directly to the results of the first and second convolutions\footnote{
We remind that Tucker-2 convolution decomposes into three consecutive smaller convolutions.} 
instead of carrying out $U_1 \odot_2 m_1$, $U_2 \odot_2 m_2$, $\mathcal{G} \odot_1 m_1 \odot_2 m_2$.  
We use the cross-entropy loss as the negative model log-likelihood.
We use Adam~\citep{DBLP:journals/corr/KingmaB14} as the optimizer of choice.
The temperature $\tau$ is exponentially decayed from $10^{-1}$ to $10^{-2}$ in the course of training. 
We discovered that \emph{hard concrete} trick~\citep{louizos2018learning}, i.e., stretching the Binary Concrete distribution and then transforming its samples with a hard-sigmoid, allows achieving better results due to inclusion of $\{0, 1\}$ into the support. 
We also found that warming up with a plain tensorized model for several epochs can help optimization, therefore, we do not apply masks multiplication at the first epochs in most of our experiments.

In Tensor Train models we do not shrink the first and the last ranks, as they equal $1$ by definition.

\section{ADDITIONAL EXPERIMENT: MARS WITH OTHER TENSORIZED MODELS}
\label{app:tucker_approx}

In this section, we demonstrate that MARS provides a unified way to select ranks in arbitrary models leveraging decomposed tensors that we name tensorized models, thus it should not be considered solely as a compression technique for neural networks. 
Namely, we consider a low-rank tensor approximation task as an illustrative example.
We emphasize that our method can be naturally extended to other tasks by substituting the given likelihood with other cost functions depending on tensor parameters.

First, we construct a $4$-dimensional tensor $\mathcal{A}$ with shape $\dims\left(\mathcal{A}\right) = (d, d, d, d), d = 8$ from a random Tucker decomposition with ranks $\bm{r} = \{r, r, r, r\}, r = 4$, i.e., 
\begin{equation*}
    \mathcal{A} = \mathcal{G}^{\mathcal{A}} \times_1 U_1^{\mathcal{A}} \dots \times_4 U_4^{\mathcal{A}},
\end{equation*}
where cores $\mathcal{G}^{\mathcal{A}} \in \mathbb{R}^{r \times r \times r \times r}, U^{\mathcal{A}}_k \in \mathbb{R}^{d \times r}, k =1\dots4$ are randomly initialized from a standard normal distribution.
Then we consider a tensorized model parameterized by cores $\bm{G}^{\mathcal{B}} = \left\{\mathcal{G}^{\mathcal{B}}, U^{\mathcal{B}}_1, \dots, U^{\mathcal{B}}_4\right\}$, where $\mathcal{G}^{\mathcal{B}} \in \mathbb{R}^{R \times R \times R \times R}, U^{\mathcal{B}}_k \in \mathbb{R}^{d \times R}, k =1\dots4, R = 8$, with the log-likelihood defined as a negative MSE between tensor 
\begin{equation*}
    \mathcal{B}\left(\bm{G}^{\mathcal{B}}\right) = \mathcal{G}^{\mathcal{B}} \times_1 U_1^{\mathcal{B}} \dots \times_4 U_4^{\mathcal{B}}
\end{equation*}
and tensor $\mathcal{A}$, i.e.,
\begin{equation*}
    \log p\left(\bm{G}^{\mathcal{B}}\right) = -\frac{1}{d^4}\left\Vert \mathcal{B}\left(\bm{G}^{\mathcal{B}}\right) - \mathcal{A} \right \Vert^2.
\end{equation*}
In other words, this model approximates the given tensor $\mathcal{A}$ with an $8$-rank Tucker decomposition.
Due to redundancy induced by the construction of $\mathcal{A}$ (GT rank is $r = 4$), this model naturally requires ranks selection.

We applied MARS with $\pi = 10^{-2}$ and $\alpha = -0.5$ to the described model.
We trained using standard gradient descent with LR $10^{-2}$ for $10^4$ epochs to achieve full convergence.
Mean results over 10 runs are the following: ranks $\bm{r} = \{4.7, 4.4, 4.6, 4.2\}$ and log-likelihood $\log p\left(\bm{G}^{\mathcal{B}}\right) = -0.027$, which is very close to the GT solution.
For comparison, a typical MSE value between two random tensors initialized according to the scheme of $\mathcal{A}$ is $\approx 300$.

Moreover, we trained a similar model with $R = r = 4$ without MARS (because there are no more extra ranks) and achieved a log-likelihood value of only $\approx -0.15$. 
We believe that more degrees of freedom in the redundant model helps MARS find a more optimal solution than even starting from the GT ranks.
It would be interesting to get a deeper theoretical insight explaining the differences in the experimental results of the direct learning methods with lower ranks and MARS.

In this synthetic experiment, we demonstrated that MARS is applicable for tensorized models other than tensorized neural networks and can be used to efficiently restore GT ranks in, e.g., low-rank tensor approximation task.
Yet, we can expect that other methods, e.g., Tucker-ARD~\citep{doi:10.1002/cem.1223}, tailored explicitly for automatic ranks determination in tensor approximation, might exhibit similarly on this simple problem. 
Further extension of MARS to other tasks along with a detailed comparison with relevant baselines is future work.

\section{ABLATION STUDY: INFLUENCE OF HYPERPARAMETER \boldmath$\pi$}
\label{app:pi_ablation}

\begin{figure}[ht]
\centering
\includegraphics[width=0.95\linewidth]{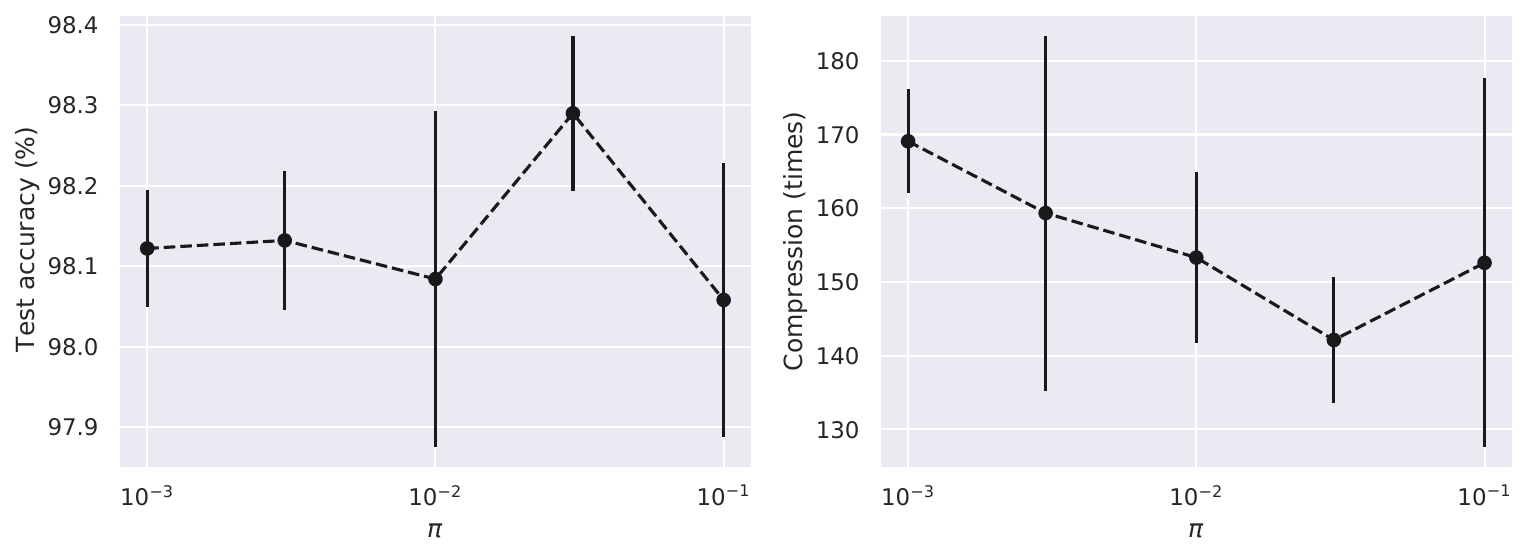}
\caption{Test accuracy (left) and compression (right) for different values of hyperparameter $\pi$. Mean $\pm$ std over 5 runs is reported for each value of $\pi$.}
\label{fig:pi_ablation}
\end{figure}

In this section, we report the results of an ablation experiment concerning the influence of the value of hyperparameter $\pi$ in MARS on the final model performance and compression.
Specifically, we repeat the 2FC-Net MNIST experiment from~\Cref{sec:mnist_2fc} using soft-mode initialization ($\alpha = -1.5$) with different values of $\pi$.
We present the results in~\Cref{fig:pi_ablation}.

As can be seen from the plots, the final test accuracies of the trained models are similar and weakly depend on the value of hyperparameter $\pi$.
At the same time, there is a tendency for compression to gradually decrease with increasing $\pi$, which fully corresponds to its role in our model.
Note also that too high values of $\pi$ can lead to inconsistent results and harm performance, since the prior term in model~\eqref{eq:model} loses its ability to promote sparse mask solutions to select optimal ranks.

In the end, we conclude that taking $\pi \approx 10^{-2}$ is a reasonable choice to achieve satisfactory compression-accuracy trade-off in most situations.

\section{LIMITATIONS AND SOCIETAL IMPACT}
\label{app:limitations}

In this section, we want to discuss the limitations and possible societal impact of our work.

We position our method, MARS, as a general and efficient way to select ranks in various tensorized models.
While we demonstrate its possible applicability to other tasks in the previous section, in our main experiments, we mostly concentrate on tensorized neural networks, since they are widely known in the community and overall recognized as a promising direction.
Further development of MARS and its evaluation on other tensor models and problems is a promising future research.

Another limitation of our work is the simplicity of the considered family of distributions over masks. 
We conjecture that more advanced variational families could put our inference method beyond a simple MAP estimate and/or further improve MARS performance.
We also find this an important direction for future studies.

While modern DNN models continue increasing the total number of learnable parameters, they require more computational resources than ever before.
This circumstance under no doubts imposes negative environmental influence.
As discussed in \Cref{app:rel_work}, tensor methods allow a significant reduction of occupied resources and energy consumption.
We hope that our method will further disseminate these methods and serve to promote environmental protection.
To the best of our knowledge, we cannot find any negative consequences from the misuse of the paper’s contribution.

\end{document}